\documentclass{article}

\usepackage[final]{nips_2017}

\usepackage[utf8]{inputenc} 
\usepackage[T1]{fontenc}    
\usepackage{hyperref}       
\usepackage{url}            
\usepackage{booktabs}       
\usepackage{amsfonts}       
\usepackage{nicefrac}       
\usepackage{microtype}      
\usepackage{graphicx}
\usepackage[english]{babel}
\usepackage{amsmath}

\title{Representation and Reinforcement Learning for Personalized Glycemic Control in Septic Patients}

\author{
  Wei-Hung Weng\\
  MIT CSAIL \\
  \texttt{ckbjimmy@mit.edu} \\  
  \And
  Mingwu Gao\\
  Philips Connected Sensing Venture \\
  \texttt{barton.gao@philips.com} \\
  \And
  Ze He\\
  Philips Research North America \\
  \texttt{ze.he@philips.com} \\
  \And
  Susu Yan\\
  Masschusetts General Hospital \\
  \texttt{syan5@mgh.harvard.edu} \\
  \And
  Peter Szolovits\\
  MIT CSAIL \\
  \texttt{psz@mit.edu} \\
}

\begin{document}

\maketitle

\begin{abstract}
Glycemic control is essential for critical care.
However, it is a challenging task because there has been no study on personalized optimal strategies for glycemic control.
This work aims to learn personalized optimal glycemic trajectories for severely ill septic patients by learning data-driven policies to identify optimal targeted blood glucose levels as a reference for clinicians.
We encoded patient states using a sparse autoencoder and adopted a reinforcement learning paradigm using policy iteration to learn the optimal policy from data.
We also estimated the expected return following the policy learned from the recorded glycemic trajectories, which yielded a function indicating the relationship between real blood glucose values and 90-day mortality rates.
This suggests that the learned optimal policy could reduce the patients' estimated 90-day mortality rate by 6.3\%, from 31\% to 24.7\%.
The result demonstrates that reinforcement learning with appropriate patient state encoding can potentially provide optimal glycemic trajectories and allow clinicians to design a personalized strategy for glycemic control in septic patients.
\end{abstract}

\section{Introduction}
Many critically ill patients have poor glycemic control, such as dysglycemia and high glycemic variability.
Several studies have investigated the correlation between the management of blood glucose levels and outcomes in critical care patients~\citep{inzucchi2006,chase2008,vandenberghe2001,finfer2009}.
Current clinical practice follows the guidelines suggested by the NICE-SUGAR trial for patients in critical care, which targets a blood glucose level of 100-180 mg/dl~\citep{finfer2009}.
However, there are significant variations in clinical conditions and physiological states among patients under critical care, including disease severity, comorbidity, and other factors that may limit clinicians' ability to perform appropriate glycemic control.
In addition, clinicians sometimes may not focus on the issue of glycemic control during patients' intensive care unit (ICU) stays.
To help clinicians better address the challenge of managing patients' glucose levels, we need a personalized glycemic control strategy that can take into account variations in patients' physiological and pathological states, such as their degree of insulin dependence and tolerance. 

Our goal of decision support for glycemic control is to target specific ranges of serum glucose that optimize outcomes for patients.  These target ranges vary based on patients' specific circumstances, and we hypothesize that the patient states, glycemic values, and patient outcomes can be modeled by a Markov decision process (MDP) whose parameters and optimal policies can be learned from data gathered from previous cases. In our approach, we do not try to recommend specific interventions such as insulin oral hypoglycemic agent administration to achieve the target goals, but focus on finding the optimal targets. Thus the ``action'' in our formulation is interpreted as choosing the best glycemic target under the circumstances, leaving the choice of agents and doses to achieve that target to the clinicians. This simplification avoids the need to model the variability of patients' glycemic responses to actual insulin doses and avoids data problems with incorrectly recorded dose timing.

In this study, we proposed and explored the reinforcement learning (RL) paradigm to learn the policy for choosing personalized optimal glycemic trajectories using retrospective data.
We then compared the prognosis of the trajectories simulated by following the optimal policy to the real trajectories.

\begin{figure}[ht]
  \caption{Study overview. Red dot lines indicate the training phase to obtain policies, and blue dashed lines indicate the testing phase to compute results.}
  \label{fig:overview}
  \centering
  \includegraphics[width=1\textwidth]{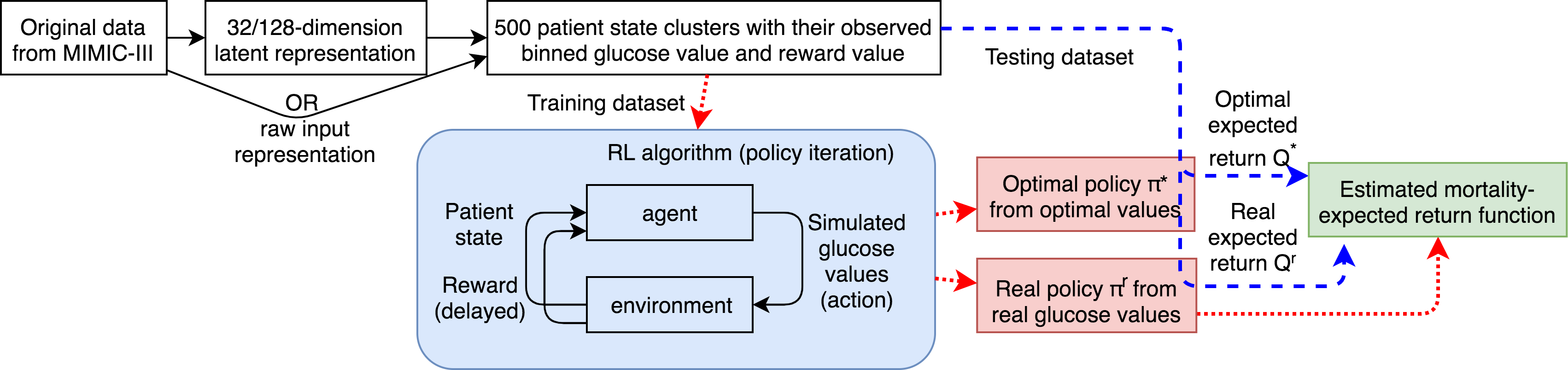}
\end{figure}

The study overview is shown in Figure~\ref{fig:overview}.
We used both direct raw features and features encoded by a sparse autoencoder to represent patient states, followed by state clustering~\citep{ng2011}.
Policy iteration was adopted to learn the optimal policy $\pi^*$ for optimal glycemic trajectories~\citep{howard1960dynamic}. We also estimated the policy $\pi^r$ from the real trajectories and obtained the estimated mortality--expected return function.
We then computed the optimal expected return $Q^*$ using optimal policy $\pi^*$, and used the function to calculate and compare the estimated mortality rate between the simulated optimal glycemic trajectories and the trajectories in the real data~\citep{komorowski2016}.

The key contributions of this work include: (a) we explored the effectiveness and validity of using RL and appropriate representations for learning the policy for optimal glycemic trajectories in different clinical and physiological states, (b) we applied the method to glycemic control for critical septic patients, and compared the RL policy-simulated results with real data. Our proposed approach may enable wider applications in personalized clinical decision making.

\section{Related work}
RL is a potential approach for sequential decision making with delayed reward or outcome~\citep{nemati2016}.
RL also has the ability to generate optimal strategies based on non-optimized training data~\citep{raghu2017}.
~\citet{shortreed2010} investigated the performance of the RL-learned optimal policy for treatment of schizophrenia.
In~\citet{nemati2016}, RL is applied to the heparin dosing problem in critical care using partially observed Markov decision process (POMDP). They used a discriminative hidden Markov model (DHMM) to estimate the patient states.
\citet{prasad2017} experimented on using off-policy learning to find the policy for mechanical ventilation administration and weaning.
~\citet{komorowski2016} and \citet{raghu2017} demonstrated that optimal strategy of sepsis treatment can be learned by RL algorithms.
~\citet{komorowski2016} applied on-policy learning and evaluated the performance difference between RL agent's actions and physicians' decisions using Q-values.
~\citet{raghu2017} worked on the same clinical problem but introduced a continuous state representation and a deep learning approach with doubly robust off-policy evaluation~\citep{jiang2015}.
Related to glycemic control, some studies utilize RL to design clinical trials and adjust clinical treatments~\citep{asoh2013, bothe2014}.
 
These previous studies focused on giving direct suggestions for clinical intervention and treatment.
To our knowledge, no studies have yet utilized the RL approach to learn better target laboratory values as references for clinical decision making, such as learning a policy to find personalized optimal glycemic targets for different patient states.

\section{Method}

\paragraph{Patient State Representation and Clustering}
Learning a good patient state representation is important to enable RL algorithms to learn appropriate policies.
We experimented with two types of feature representations: raw interpretable clinical features and the feature representation generated by a sparse autoencoder~\citep{ng2011}.
We designed the neural network model with one encoder and decoder, and extracted the latent representation as 32-dimensional states.
The weights of the sparse autoencoder were computed by backpropagation to minimize the following loss function: $\mathcal{L}_{\mathit{SparseAE}} = \mathcal{L}_{\mathit{AE}} + \beta \cdot \Sigma_{i=1}^{n} D \cdot \log \frac{D}{h_i} + (1-D) \cdot \frac{1-D}{1-h_i}$, where $\mathcal{L}_{\mathit{AE}}$ is the reconstruction loss of the autoencoder, $D$ is the sparsity distance, $h$ is the latent representation, $n$ is the dimension of the latent representation, and $\beta$ is a hyperparameter for sparsity.
After we generated the state representation, we used the k-means clustering algorithm to categorize millions of patient states into 500 clusters such that similar clinical states can collapse into the same cluster.
%
%
%

\paragraph{Reinforcement Learning and Evaluation}
The policy iteration algorithm uses dynamic programming and follows the Bellman expectation equation to find the optimal policy $\pi^{*}$ through two steps~\citep{howard1960dynamic,bellman1957dynamic}, iterative policy evaluation and greedy policy improvement. 
Given a current policy $\pi$, reward function $\mathcal{R}$, transition function $\mathcal{P}$, the current state $s$ and the next state $s'$ after taking action $a$, the threshold for iteration $\epsilon=10^{-4}$ and the discount factor $\gamma=0.9$, we iteratively updated state-value functions $v_k$ of all states: $v_{k+1}(s) = \sum_{a \in \mathcal{A}} \pi(a|s) (\mathcal{R}_s^a + \gamma \mathcal{P}_{ss'}^{a} v_k(s'))$. Then we returned $v_{\pi} = v_{k+1}$ once $|v_{k+1} - v_{k}| < \epsilon$. 
Next we used the value function $v_{\pi}$ to obtain a better policy by greedy policy improvement using the Bellman expectation equation as follows: $Q^{\pi}(s, a) = \mathcal{R}_s^a + \gamma \sum_{s' \in \mathcal{S}} \mathcal{P}_{ss'}^{a} v^{\pi}(s')$, and $\pi'(s) = \arg\max_{a \in \mathcal{A}} Q^{\pi}(s, a)$. 
We used the new policy $\pi'$ to update new value functions until $\pi$ converges to the optimal policy $\pi^*$. 
The optimal expected return $Q^* = Q^{\pi^{*}}$. 
%
The algorithm was adopted to learn the optimal policy $\pi^*$, and the real policy $\pi^r$ from real trajectories, where we limited the action space of each state to only the one with the highest probability in $\mathcal{P}$ instead of exploring all possible actions. 
$\pi^r$ and real mortality rate were used to obtain the estimated mortality--expected return function, which reveals the relationship between expected return and the estimated 90-day mortality rate.
We used the function to compute and compare the estimated mortality rate of real and optimal glycemic trajectories obtain by $\pi^r$ and $\pi^*$.

\section{Experiments}

\paragraph{Dataset and Patient Cohort}
We used a large, publicly available ICU database, Medical Information Mart for Intensive Care III (MIMIC-III) version 1.4~\citep{johnson2016} for the study.
MIMIC-III includes all patients admitted to an ICU at the tertiary medical center, Beth Israel Deaconess Medical Center (BIDMC) in Boston, MA from 2001 to 2012.
It contains all ICU administrative data such as service and diagnosis codes, clinical data such as lab and bedside measurements, as well as the death time (in case of mortality) from the social security database.

Sepsis-3 criteria were adopted to identify patients with sepsis~\citep{singer2016}.
All patients aged less than 18 years or whose Sequential Organ Failure Assessment (SOFA) score was less than two were excluded from the study cohort.
For the subjects with multiple ICU admissions, only the first ICU admission was selected in order to ensure the independence of observations.
We specifically separated the patients by correct diabetic status because it is critical to understanding the blood glucose trajectory.
We categorized the patients into diabetic and non-diabetic groups according to their (1) ICD-9 coding for diabetes (249.* and 250.*), (2) pre-admission glycated hemoglobin (HbA1c) level with the threshold of 7.0\%, (3) admission medications, and (4) the diagnosis of diabetes in the free text medical history.
We removed patients with more than 10\% missing values of covariates, and the final cohort comprised 5,565 patients.
Only arterial and venous glucose measurements were considered due to the issues of accuracy of other measurements~\citep{critchell2007accuracy}.
%
%
%

\paragraph{Settings of Reinforcement Learning}
To decrease the potential for confounding, we considered the following groups of confounders and encoded them as covariates into the patient states.
Patient level variables: age, gender, admitted ICU unit, SOFA and Elixhauser comorbidity index on admission, utilization of mechanical ventilator, endotracheal intubation or vasopressor.
Blood glucose related variables: glycated hemoglobin (HbA1c) higher or less than 7 before admission, hyperglycemia and hypoglycemia events occurred, the first glucose level after admission, and the diabetic status of the patient.
Periodic vital signs, such as systolic and diastolic blood pressure, respiratory rate, temperature, heart rate, and SpO\textsubscript{2} and Glasgow coma scale (eyes, verbal, motor), as well as laboratory values, such as arterial blood gas (PO\textsubscript{2}, PCO\textsubscript{2}, pH value, total CO\textsubscript{2}, anion gap), albumin, bicarbonate, calcium, sodium, potassium, chloride, lactate, creatinine, blood urea nitrogen (BUN), partial thromboplastin time (PTT), international normalized ratio (INR), CO\textsubscript{2}, total bilirubin, hemoglobin, hematocrit and white blood cell count.
Data were collected at one hour intervals.
We imputed missing values by linear interpolation of the existing values, and used piecewise constant interpolation if the first or last observation is missing.
%
%
%
All variables were normalized into a zero to one scale for the state encoding.

We used the outcome of interest, 90-day mortality status, as the reward since patients' survival is clinicians' major goal for critical care.
A numeric value of +100 was assigned to the end state for patients who survived 90 days after their admission, and -100 as a penalty for those who were deceased before 90 days after their admission. 
The actions were defined by the discretized blood glucose levels, which were calculated from categorizing continuous glucose levels into 11 bins. 

\begin{figure}[ht]
  \caption{Estimated mortality rate versus expected return computed by the learned policies. (left) Raw feature representation, (right) 32-dimension sparse autoencoder representation}
  \label{fig:curve}
  \centering
  \includegraphics[width=0.67\textwidth]{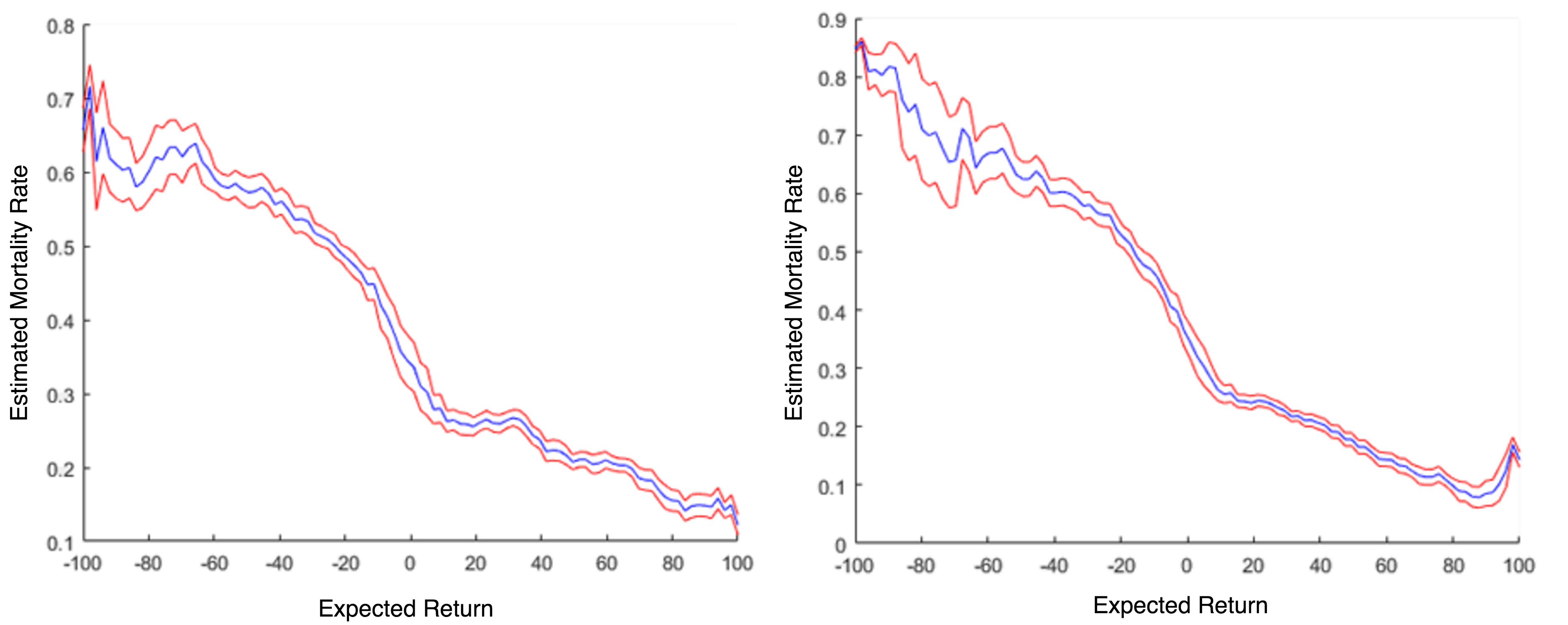}
\end{figure}

\paragraph{Results}
In Figure~\ref{fig:curve}, the expected return, which is the rescaled $Q^*$ learned by the optimal policy $\pi^*$, is negatively correlated with mortality rate with high correlation.
This indicates that the learned expected return reflects the real patient status well.
We computed the empirically estimated mortality rate of the real glycemic trajectory using the mortality--expected return function acquired from raw and encoded patient state representations.
The average expected 90-day mortality rate of the testing dataset is 31.00\% using the raw feature representation, and 31.08\% using the sparse autoencoder latent representation. 
Both are close to the mortality rate calculated from the real data (31.17\%).

Using the raw features to learn policy from the real glycemic trajectory, the mean expected return was 10.04, which was consistent with an estimated mortality rate of 31.00\%.
The optimal policy-simulated glycemic trajectories had a mean expected return of 36.42, which corresponded to an estimated mortality rate of 27.29\%.
Using the latent representations learned by the sparse autoencoder further reduced the estimated mortality rate to 24.75\%.
The best optimal policy can potentially reduce the estimated mortality rate by 6.3\% if we choose the appropriate representations (Table~\ref{table1}).

\begin{table}[ht]
  \scriptsize
  \caption{Estimated mortality rate of glycemic trajectories simulated by real/optimal policies}
  \label{table1}
  \centering
  \begin{tabular}{lllll}
    \toprule
    & Real policy & & Optimal policy & \\
    \midrule
    Representation & Expected return & Estimated mortality & Expected return & Estimated mortality \\
    \midrule
    Raw features & 10.04 & 31.00\% & 36.42 & 27.29\% \\
    Sparse autoencoder & 8.75 & 31.08\% & 32.49 & 24.75\% \\
    \bottomrule
  \end{tabular}
\end{table}

\section{Conclusions and Future Works}
We utilized the RL algorithm with representation learning to learn the personalized optimal policy for better predicting glycemic targets from retrospective data. 
As decision support, this may reduce the mortality rate of septic patients if clinician-chosen dosages can actually achieve the target glucose levels chosen by the policy, and thus potentially assist clinicians to optimize the real-time treatment strategy by providing more accurate treatment goals.
Future works include applying a continuous state space approach, different evaluation methods such as doubly robust evaluation, as well as applying the method to different clinical decision making problems. 

\bibliography{nips_2017_rl}

\begin{thebibliography}{18}
\providecommand{\natexlab}[1]{#1}
\providecommand{\url}[1]{\texttt{#1}}
\expandafter\ifx\csname urlstyle\endcsname\relax
  \providecommand{\doi}[1]{doi: #1}\else
  \providecommand{\doi}{doi: \begingroup \urlstyle{rm}\Url}\fi

\bibitem[Asoh et~al.(2013)Asoh, Shiro, Akaho, Kamishima, Hasida, Aramaki, and
  Kohro]{asoh2013}
H.~Asoh, M.~Shiro, S.~Akaho, T.~Kamishima, K.~Hasida, E.~Aramaki, and T.~Kohro.
\newblock An application of inverse reinforcement learning to medical records
  of diabetes treatment.
\newblock \emph{ECMLPKDD2013 Workshop on Reinforcement Learning with
  Generalized Feedback}, 2013.

\bibitem[Bellman(1957)]{bellman1957dynamic}
R.~Bellman.
\newblock \emph{Dynamic Programming.}
\newblock Princeton University Press, Princeton, NJ, 1957.

\bibitem[Bothe et~al.(2014)Bothe, Dickens, Reichel, Tellmann, Ellger, Westphal,
  and Faisal]{bothe2014}
M.~K. Bothe, L.~Dickens, K.~Reichel, A.~Tellmann, B.~Ellger, M.~Westphal, and
  A.~A. Faisal.
\newblock The use of reinforcement learning algorithms to meet the challenges
  of an artificial pancreas.
\newblock \emph{Expert Review of Medical Devices}, 10\penalty0 (5):\penalty0
  661--673, 2014.

\bibitem[Chase et~al.(2008)Chase, Shaw, Le~Compte, Lonergan, Willacy, Wong,
  Lin, Lotz, Lee, and Hann]{chase2008}
J.~G. Chase, G.~Shaw, A.~Le~Compte, T.~Lonergan, M.~Willacy, X.-W. Wong,
  J.~Lin, T.~Lotz, D.~Lee, and C.~Hann.
\newblock Implementation and evaluation of the {SPRINT} protocol for tight
  glycaemic control in critically ill patients: a clinical practice change.
\newblock \emph{Critical Care}, 12\penalty0 (2):\penalty0 R49, 2008.

\bibitem[Critchell et~al.(2007)Critchell, Savarese, Callahan, Aboud, Jabbour,
  and Marik]{critchell2007accuracy}
C.~D. Critchell, V.~Savarese, A.~Callahan, C.~Aboud, S.~Jabbour, and P.~Marik.
\newblock Accuracy of bedside capillary blood glucose measurements in
  critically ill patients.
\newblock \emph{Intensive care medicine}, 33\penalty0 (12):\penalty0
  2079--2084, 2007.

\bibitem[Howard(1960)]{howard1960dynamic}
R.~A. Howard.
\newblock \emph{Dynamic Programming and Markov Processes.}
\newblock MIT Press, Cambridge, MA, 1960.

\bibitem[Inzucchi(2006)]{inzucchi2006}
S.~E. Inzucchi.
\newblock Management of hyperglycemia in the hospital setting.
\newblock \emph{New England Journal of Medicine}, 355\penalty0 (18):\penalty0
  1903--1911, 2006.

\bibitem[Jiang and Li(2015)]{jiang2015}
N.~Jiang and L.~Li.
\newblock Doubly robust off-policy evaluation for reinforcement learning.
\newblock \emph{arXiv preprint arXiv:1511.03722}, 2015.

\bibitem[Johnson et~al.(2016)Johnson, Pollard, Shen, Lehman, Feng, Ghassemi,
  Moody, Szolovits, Celi, and Mark]{johnson2016}
A.~E. Johnson, T.~J. Pollard, L.~Shen, L.-w.~H. Lehman, M.~Feng, M.~Ghassemi,
  B.~Moody, P.~Szolovits, L.~A. Celi, and R.~G. Mark.
\newblock {MIMIC}-{III}, a freely accessible critical care database.
\newblock \emph{Scientific Data}, 3:\penalty0 160035, 2016.

\bibitem[Komorowski et~al.(2016)Komorowski, Gordon, Celi, and
  Faisal]{komorowski2016}
M.~Komorowski, A.~Gordon, L.~A. Celi, and A.~Faisal.
\newblock A markov decision process to suggest optimal treatment of severe
  infections in intensive care.
\newblock \emph{Neural Information Processing Systems Workshop on Machine
  Learning for Health}, 2016.

\bibitem[Nemati et~al.(2016)Nemati, Ghassemi, and Clifford]{nemati2016}
S.~Nemati, M.~M. Ghassemi, and G.~D. Clifford.
\newblock Optimal medication dosing from suboptimal clinical examples: A deep
  reinforcement learning approach.
\newblock \emph{Conference proceedings: Annual International Conference of the
  IEEE Engineering in Medicine and Biology Society. IEEE Engineering in
  Medicine and Biology Society. Annual Conference}, pages 2978--2981, 2016.

\bibitem[Ng(2011)]{ng2011}
A.~Y. Ng.
\newblock Sparse autoencoder.
\newblock 2011.
\newblock URL
  \url{{https://web.stanford.edu/class/archive/cs/cs294a/cs294a.1104/sparseAutoencoder.pdf}}.

\bibitem[Prasad et~al.(2017)Prasad, Cheng, Chivers, Draugelis, and
  Engelhardt]{prasad2017}
N.~Prasad, L.-F. Cheng, C.~Chivers, M.~Draugelis, and B.~E. Engelhardt.
\newblock A reinforcement learning approach to weaning of mechanical
  ventilation in intensive care units.
\newblock \emph{arXiv preprint arXiv:1704.06300v1}, 2017.

\bibitem[Raghu et~al.(2017)Raghu, Komorowski, Celi, Szolovits, and
  Ghassemi]{raghu2017}
A.~Raghu, M.~Komorowski, L.~A. Celi, P.~Szolovits, and M.~Ghassemi.
\newblock Continuous state-space models for optimal sepsis treatment - a deep
  reinforcement learning approach.
\newblock \emph{arXiv preprint arXiv:1705.08422}, 2017.

\bibitem[Shortreed et~al.(2010)Shortreed, Laber, Lizotte, Stroup, Pineau, and
  Murphy]{shortreed2010}
S.~M. Shortreed, E.~Laber, D.~J. Lizotte, T.~S. Stroup, J.~Pineau, and S.~A.
  Murphy.
\newblock Informing sequential clinical decision-making through~reinforcement
  learning: an empirical study.
\newblock \emph{Machine Learning}, 84\penalty0 (1-2):\penalty0 109--136, 2010.

\bibitem[Singer et~al.(2016)Singer, Deutschman, Seymour, Shankar-Hari, Annane,
  Bauer, and et~al]{singer2016}
M.~Singer, C.~S. Deutschman, C.~W. Seymour, M.~Shankar-Hari, D.~Annane,
  M.~Bauer, and et~al.
\newblock The third international consensus definitions for sepsis and septic
  shock (sepsis-3).
\newblock \emph{JAMA}, 315\penalty0 (8):\penalty0 801--810, 2016.

\bibitem[{The NICE-SUGAR Study Investigators}(2009)]{finfer2009}
{The NICE-SUGAR Study Investigators}.
\newblock Intensive versus conventional glucose control in critically ill
  patients.
\newblock \emph{New England Journal of Medicine}, 360\penalty0 (13):\penalty0
  1283--1297, 2009.

\bibitem[Van~den Berghe et~al.(2001)Van~den Berghe, Wouters, Weekers, Verwaest,
  Bruyninckx, Schetz, Vlasselaers, Ferdinande, Lauwers, and
  Bouillon]{vandenberghe2001}
G.~Van~den Berghe, P.~Wouters, F.~Weekers, C.~Verwaest, F.~Bruyninckx,
  M.~Schetz, D.~Vlasselaers, P.~Ferdinande, P.~Lauwers, and R.~Bouillon.
\newblock Intensive insulin therapy in critically ill patients.
\newblock \emph{New England Journal of Medicine}, 345\penalty0 (19):\penalty0
  1359--1367, 2001.

\end{thebibliography}
\bibliographystyle{abbrvnat}

\end{document}